\documentclass[letterpaper, 10 pt, conference]{ieeeconf}  

\IEEEoverridecommandlockouts                              

\usepackage{graphicx} 
\usepackage{epsfig} 
\usepackage{times} 
\usepackage{amsmath} 
\usepackage{amssymb}  
\usepackage{bm}
\usepackage{subfig}
\usepackage{url}
\usepackage{color}
\usepackage{cite}
\newtheorem{remark}{Remark}
\usepackage{rotating}
\usepackage{afterpage}

\usepackage[ruled]{algorithm2e}

\SetCommentSty{mycommfont}

\usepackage{algpseudocode}
\algnewcommand\algorithmicforeach{\textbf{for each}}
\algdef{S}[FOR]{ForEach}[1]{\algorithmicforeach\ #1\ \algorithmicdo}

\DeclareMathOperator*{\argmin}{argmin}
\title{\LARGE \bf
	An observable time series based SLAM algorithm for deforming environment}

\author{Jingwei Song$^{}$, Liang Zhao$^{}$, Shoudong Huang$^{}$ and Gamini Dissanayake$^{}$
	\thanks{$^{}$All the authors are from Centre for Autonomous Systems, University of Technology, Sydney, P.O. Box 123, Broadway, NSW 2007, Australia}%
	\thanks{Email: jingwei.song; @student.uts.edu.au, \{Liang.Zhao; Shoudong.Huang;
		Gamini.Dissanayake\}@uts.edu.au} 
}

\begin{document}

	\maketitle
	\thispagestyle{empty}
	\pagestyle{empty}
	
	\begin{abstract}		In this paper, we study the back-end of simultaneous localization and mapping (SLAM) problem in deforming environment, where robot localizes itself and tracks multiple non-rigid soft surface using its onboard sensor measurements. An elaborate analysis is conducted on conventional deformation modelling method, Embedded Deformation
	(ED) graph. We demonstrate and prove that the ED graph widely used in such scenarios is unobservable and leads to multiple solutions unless suitable priors are provided. Example as well as theoretical prove are provided to show the ambiguity of ED graph and camera pose. In modelling non-rigid scenario with ED graph, motion priors of the deforming environment is essential to separate robot pose and deforming environment. The conclusion can be extrapolated to any free form deformation formulation. In solving the observability, this research proposes a preliminary deformable SLAM approach
to estimate robot pose in complex environments that exhibits regular motion.
A strategy that approximates deformed shape using a linear combination of several previous shapes is proposed to
avoid the ambiguity in robot movement and rigid and
non-rigid motions of the environment. Fisher information matrix rank analysis with a base case is discussed to prove the effectiveness. 
Moreover, the proposed algorithm is validated extensively on Monte Carlo simulations and real experiments. It is demonstrated that the new algorithm significantly outperforms
conventional rigid SLAM
and ED based SLAM especially in scenarios where there is large deformation. 
		
	\end{abstract}

	\section{INTRODUCTION}
	
	SLAM, simultaneous localization and mapping, is a technique applied in robotics for pose estimation and environment mapping. Considerable research efforts have been devoted to solve robot navigation environment mapping in rigid scenario. Robust and accurate SLAM systems are now open-source including PTAM \cite{klein2007parallel}, Kinectfusion \cite{newcombe2011kinectfusion}, LSD-SLAM \cite{engel2014lsd}, ORB-SLAM \cite{mur2015orb} and Bundlefusion \cite{dai2017bundlefusion}. The algorithms and implementations are well designed and adjusted, making these approaches widely applied in indoor and outdoor real-time localization and mapping. The SLAM algorithms are robustness against noises and efficiency in practice; they are now commercialized and widely applied in applications like Virtual reality and Augmented reality. \par
	
	Inspired by the achievements in rigid scenario, interests are now shifted towards exploiting the feasibility of applying SLAM to other special scenarios, like doing SLAM with unconventional sensors or in changing environment. Nevertheless, traditional techniques like pose graph, image processing, close looping, out-lier detection are all valuable in processing deformable SLAM problems. In this article, we focus on one hot topic of 3D reconstruction and SLAM field, that is mapping and tracking of surfaces with deformation. There are three major solutions in deforming scenario: modified rigid SLAM algorithms, embedded deformation (ED) graph based algorithms and Finite Element Method (FEM) based approaches.\par 
	Many researches adopt aforementioned rigid SLAM algorithms with modifications. \cite{grasa2011ekf} and \cite{lin2013simultaneous} employ conventional feature based SLAM including extended Kalman filter (EKF) SLAM and PTAM. Thresholds are exerted to separate rigid and non-rigid feature points. \cite{mahmoud2016orbslam,mahmoud2017slam,turan2017non} adopt ORB-SLAM \cite{mur2015orb} with deliberately adjusted parameters for mapping and localizing monoscope in surgical scenario. The results they present demonstrate ORB-SLAM can also robustly localize camera in deformable environment. Recently, \cite{chen2018slam} and \cite{mahmoud2019live} apply ORB-SLAM in augmented reality enhanced surgical systems. The key modification of rigid SLAM system in non-rigid scenario is to arbitrarily choose strategy to determine rigid feature points for localization. Similarly, those redeemed non-rigid features are discarded in localization steps. Though enhanced with numerous technical modules, the basic state transition function and observation function of rigid SLAM algorithms are the same. Therefore, feature motions models of these SLAM algorithms are identical to classical SLAM systems, but with additional presumption that part of environment is rigid. Other than the unreliable rigid feature identification strategy, all delicate modules of ORB-SLAM are exploited and benefited by these methods for enhancing robustness and accuracy.\par 
	
	Different from rigid feature assumption, embedded deformation (ED) graph based algorithm is priori free, and introduces deformation field alongside rigid transformation. Acknowledging all features are skeptical to deformation, a warping field is introduced to model deformation based on assumption of 'As-rigid-as-possible' proposed by \cite{sorkine2007rigid}. The difference between SLAM in deformable environment and in dynamic environment \cite{saputra2018visual} are: ED graph assumes the motions are continuous in space while dynamic SLAM assumes motions are discrete and space independent.With ED graph, \cite{newcombe2015dynamicfusion,innmann2016volumedeform,guo2017real} achieve real-time non-rigid model deformation and incremental reconstruction. Meanwhile, \cite{dou2016fusion4d} proposes a complete system consisting multiple RGBD cameras as a substitution so that it builds a real-time colored, fast moving and close-loop model. These template-free techniques are able to process slow motion without occlusions because the sensor used is a single depth camera. Different from previous human body reconstruction within a volume, \cite{song2018dynamic}\cite{song2018mis} propose MIS-SLAM: a complete real-time large scale dense deformable SLAM system with stereoscope in Minimal Invasive Surgery based on heterogeneous computing. Similar works can also be found in \cite{turan2017non}. The stereo vision based system can localize the scope while build a dense deformable model. \par 
	
	Similar to ED graph, FEM is also integrated into SLAM system. FEM presents model by discreting a geometry into elements defined by 3D locations of nodes. Nodes and their connecting edges form a net and their displacements are assumed to be the deformed shape. Deformation is described by stiffness parameters controlling behaviors of nodes sharing same edge. Some works have reported its effectiveness \cite{agudo2011fem}\cite{agudo20123d}. No work demonstrates incrementally building finite element net and no complete implementations have presented the efficiency it solves dense deformable SLAM problem. \par

	While there are numerous SLAM implementations in deformable scenario, no analysis is reported on the separability, or defined as observability in classical control theory, of robot motion (rigid transformation) and environment deformation (non-rigid deformation) in deformable SLAM. In the field of control, observability of system is defined of the ability to fully and uniquely recover the system state, from a finite number of observations of its outputs and the knowledge of its control inputs \cite{bar2004estimation}. When ED graph is applied in non-rigid SLAM, the results of 3D reconstructions are intuitive, and many works ignore the accuracy of rigid pose. The appealing reconstructed geometry is a mixed result of pose tracking and environment deformation estimating. However, pose estimation is crucial in SLAM problem and we therefore focus on this topic. Particularly, the question is `Is global pose of robot observable in deformable environment unique?'. If the answer is no, then `How can we enable observability of pose in a deformable environment?'.\par 
	
	In this paper, we extensively discuss the observability in the ED graph based SLAM. There are four major contributions: (1) A counterexample is provided when analyzing ED graph based visual SLAM system in deformable environment. We clearly demonstrate that the global pose of robot can be embedded into environment deformation formulation which is not separable. (2) We theoretically prove the above conclusion by analyzing the rank of Fisher information matrix. (3) We propose an innovative back-end SLAM system which can efficiently calculate accurate pose as well as the deforming environment. (4) A prove is also provided to validate that the time series formulation is observable. The proposed time series method is inspired by Fourier Transformation. We introduce a priori that theoretically deformation is a mixture of base shapes. Typical deformations this method are suitable to handle include heartbeat, breathe, periodic body movement. Other deformation can also, to some extend, be approximated by several historical basis shapes with rigid movement. The proposed time series method explicitly enforces correct observability constraints to overcome rigid pose mixing with non-rigid deformation field. The result is compared with conventional rigid SLAM and embedded deformation formulation. \par 
	

	\section{Observability Analysis of ED based SLAM}
	\label{section:observability}
	Conventional rigid SLAM formulation fails to formulate the movement of features and attributes to underfitting. Recently, ED graph has entered the vernacular of SLAM community. Specifically, ED based approach \cite{newcombe2015dynamicfusion,innmann2016volumedeform,guo2017real,dou2016fusion4d} is widely used to define the deformation of environment. In this section, we present the basic form and the corresponding matrix formulation of ED graph. Based on the matrix formulation, observability of ED graph formulation is analyzed with an example as well as theoretical prove. Surprisingly, the rotation and translation of robot can be precisely mixed with ED graph. Based on the analysis, we conclude that the global pose and local deformation cannot be accurately estimated unless prior environment motion information is available. In next section, we propose a new time series based SLAM algorithm for better localization of the robot assuming that feature behaves in a mixture of historical trajectory.\par 
	
	Aiming at proving the ED based formulation is not observable, we follow the authentic observability definition \cite{yip1981solvability} by showing the robot pose with ED graph formulation is not solvable, that is the existence of multiple optimal solutions for the ED formulation. When there is no adequate information to uniquely obtain the solution, we call the problem is unobservable \cite{yip1981solvability}. Intuitively, the definition demonstrates the close connection within two intertwined variables, showing the underlying reason for unobservable. Moreover, we also follow another classical way of testing system observability \cite{bar2004estimation}\cite{thrun2005probabilistic} by presenting a theoretical prove based on information matrix analysis. \par

	\subsection{ED based deformation formulation}
	In ED based deformation formulation, deformation is expressed by weighted average of locally rigid rotation and transformation defined by deformation nodes, which are sparsely and evenly scattered in space. Each node j possess affine matrix $\mathbf{A}_j$ $\in\mathbb{R}^{3\times3}$ and a translation vector $\mathbf{t}_j$ $\in\mathbb{R}^3$. For a single feature point, it is transformed to its target position by several nearest ED nodes. In addition to deformation parameters, ED node $j$ also has a spatial coordinate ${\mathbf{g}_j}$ $\in\mathbb{R}^3$, defining the location of node $j$ and is independent on robot and features. For any vertex $\mathbf{v}_i$ in robot coordinate, the new position ${\tilde{\mathbf{v}}_i}$ is warped and transformed by ED nodes follows:\par
	\begin{equation}
	\tilde{\mathbf{v}}_i=\mathbf{R}_c{\sum_{j=1}^k w_j(\mathbf{v}_i)[\mathbf{A}_j(\mathbf{v}_i-\mathbf{g}_j)+\mathbf{g}_j+\mathbf{t}_j]}+\mathbf{T}_c,
	\label{TransformationFomulation}
	\end{equation}
	where $k$ is the number of neighboring nodes. $w_j(\mathbf{v}_i)$ is associated weight for transforming $\mathbf{v}_i$ exerted by one of the neighboring ED node. Additionally, $\mathbf{R}_c$ and $\mathbf{T}_c$ denote the relative rigid rotation and translation and is the same as the robot pose in rigid SLAM scenario. We confine the number of nearest nodes by defining the weight in Eq. (\ref{eq_weight}). For a single model point $\tilde{v}_i$ in robot coordinate, the weight of deforming the point from node $j$ is set as:
	\begin{equation}
	\label{eq_weight}
	w_j(\tilde{\mathbf{v}}_i)=(1-||\mathbf{v}_i-\mathbf{g}_j||/d_{max})
	\end{equation}
	where $d_{max}$ is the maximum distance of the vertex to $k + 1$ nearest ED node.\par 	
	To regularize the behavior of ED graphs, two terms \textbf{Rotation} and \textbf{Regularization} are introduced.\par 
	
	\textbf{Rotation}. $E_{rot}$ sums the rotation error of all the matrix $\mathbf{A}$ in the following form:
	\begin{equation}
	\label{Erot}
	E_{rot}=\sum_{j=1}^m Rot(\mathbf{A}_j)
	\end{equation}
	\begin{equation}
	\begin{aligned}
	Rot(\mathbf{A})=(\mathbf{c_1}\cdot\mathbf{c_2})^2+(\mathbf{c_1}\cdot\mathbf{c_3})^2+(\mathbf{c_2}\cdot\mathbf{c_3})^2+\\
	(\mathbf{c_1}\cdot\mathbf{c_1}-1)^2+(\mathbf{c_2}\cdot\mathbf{c_2}-1)^2+(\mathbf{c_3}\cdot\mathbf{c_3}-1)^2
	\end{aligned}
	\end{equation}
	where $\mathbf{c_1}$, $\mathbf{c_2}$ and $\mathbf{c_3}$ are the column vectors of the matrix $\mathbf{A}$.
	
	\newcounter{mytempeqncnt}
	\begin{figure*}[!t]
		\setcounter{equation}{5}
		\normalsize
		\begin{equation}
		\label{MatrixM}
		\underset{3m\times n}{\mathbf{M}} =
		\begin{bmatrix}
		\vdots&\vdots&\vdots&\vdots&\vdots\\
		\omega_{i=\mathbb{N}(1,1)}*(\mathbf{v}_1-\mathbf{g}_{i=\mathbb{N}(1,1)})&\cdots&\cdots&\cdots&\cdots\\
		\vdots&\vdots&\vdots&\vdots&\omega_{i=\mathbb{N}(n,1)}*(\mathbf{v}_n-\mathbf{g}_{i=\mathbb{N}(n,1)})\\
		\omega_{i=\mathbb{N}(1,2)}*(\mathbf{v}_1-\mathbf{g}_{i=\mathbb{N}(1,2)})&\cdots&\cdots&\cdots&\omega_{i=\mathbb{N}(n,2)}*(\mathbf{v}_n-\mathbf{g}_{i=\mathbb{N}(n,2)})\\
		\vdots&\vdots&\vdots&\vdots&\omega_{i=\mathbb{N}(n,3)}*(\mathbf{v}_n-\mathbf{g}_{i=\mathbb{N}(n,3)})\\
		\omega_{i=\mathbb{N}(1,3)}*(\mathbf{v}_1-\mathbf{g}_{i=\mathbb{N}(1,3)})&\cdots&\cdots&\cdots&\cdots\\
		\vdots&\vdots&\vdots&\vdots&\vdots\\
		\omega_{i=\mathbb{N}(1,4)}*(\mathbf{v}_1-\mathbf{g}_{i=\mathbb{N}(1,4)})&\cdots&\cdots&\cdots&\omega_{i=\mathbb{N}(n,4)}*(\mathbf{v}_n-\mathbf{g}_{i=\mathbb{N}(n,4)})\\
		\vdots&\vdots&\vdots&\vdots&\vdots\\
		\end{bmatrix}
		\end{equation}
		\setcounter{equation}{\value{mytempeqncnt}}
		\vspace*{4pt}
	\end{figure*}	
	
	\textbf{Regularization}. The goal is to prevent divergence of the neighboring nodes exerts on the overlapping space. For details, please refer to \cite{song2018dynamic}. \par
	\begin{equation}
	E_{reg}=\sum_{j=1}^m\sum_{k\in{\mathbb{N}(j)}} \alpha_{jk}||\mathbf{A}_j(\mathbf{g}_k-\mathbf{g}_j)+\mathbf{g}_j+\mathbf{t}_j-(\mathbf{g}_k+\mathbf{t}_k)||^2
	\label{RegulationConstraint}
	\end{equation}
	where $\alpha_{jk}$ is the overlap influence of the two ED nodes. We follow \cite{sumner2007embedded} by uniformly setting $\alpha_{jk}$ to 1. $\mathbb{N}(j)$ is the set of all neighboring nodes to the node $j$. \par
	
	The general form for ED graph problem is by minimizing the objective function:
	\begin{equation}
	\argmin\limits_{\mathbf{A}_1,\mathbf{t}_1...\mathbf{A}_m,\mathbf{t}_m,\mathbf{R}_c,\mathbf{T}_c} w_{rot}E_{rot}+w_{reg} E_{reg}+w_{data} E_{data},
	\label{Eq:energyfunction}
	\end{equation}
	where $w_{rot}$, $w_{reg}$ and $w_{data}$ are hyper-parameters. $m$ is the number of nodes in ED graph. $E_{data}$ defines an arbitrary distance from original model points to target points, planar or surface.\par  
	
	\subsection{Matrix presentation of ED graphs}
	The process of ED based deformation is within the interaction of nodes and points; the general form of whole model deformation is difficult to be expressed in the form of Eq. (\ref{TransformationFomulation}). For simplicity of notations and proof, we employ a matrix expression of process model based on Eq. (\ref{TransformationFomulation}). We first define ${\mathbf{P}}=[\mathbf{v}_1...\mathbf{v}_n]$ as the original point cloud and ${\tilde{\mathbf{P}}}=[\tilde{\mathbf{v}}_1...\tilde{\mathbf{v}}_n]$ to be the transformed point cloud. Based on definition of deformation (Eq. (\ref{TransformationFomulation})), each single point $\mathbf{v}_i$ is deformed by its $k=4$ nearest nodes and the weight is measured by the pair-wise distance. After combining all model points, we simplify this process with two influential matrices (in Eq. (\ref{MatrixM}) and Eq. (7))
	\begin{equation}	
	\label{MatrixC}
	\setcounter{equation}{7}
	\underset{m\times n}{\mathbf{C}} =
	\begin{bmatrix}
	\vdots\\
	\omega_{i=\mathbb{N}(1,1)}&\cdots&\cdots&\cdots&\cdots\\
	\vdots&\cdots&\cdots&\cdots&\omega_{i=\mathbb{N}(n,1)}\\
	\omega_{i=\mathbb{N}(1,2)}&\cdots&\cdots&\cdots&\omega_{i=\mathbb{N}(n,2)}\\
	\vdots&\cdots&\cdots&\cdots&\omega_{i=\mathbb{N}(n,3)}\\
	\omega_{i=\mathbb{N}(1,3)}&\cdots&\cdots&\cdots&\cdots\\
	\vdots&\cdots&\cdots&\cdots&\cdots\\
	\omega_{i=\mathbb{N}(1,4)}&\cdots&\cdots&\cdots&\omega_{i=\mathbb{N}(n,4)}\\
	\vdots&\cdots&\cdots&\cdots&\cdots\\
	\end{bmatrix},
	\end{equation}

	\noindent where $\mathbb{N}(i,j)(j\le 4)$ is the $j$-th neighboring node to point $\mathbf{v}_i$. Note that the element positions of weight are dependent on the topology of points-to-nodes and is not regularly arranged. Each row represents the number of points connected to this node and each column denotes how many nodes are connected to the point. Therefore, the positions of non-zero elements in each row are dependent on the neighboring node in each column. The unchanged is that each column only has 4 elements (neighbor nodes) and the sum is 1 (sum of weights).\par 
	We also define two matrices of deformation graphs relating to $\mathbf{A}_i$ and $\mathbf{t}_i$:
	\begin{equation}
	\label{MatrixA}
	{\mathbf{\Lambda}} =
	\left(
	\begin{array}{ccc}
{\mathbf{A}_1}&\cdots&{\mathbf{A}_m}\\
	\end{array}
	\right)^T
	\end{equation}
	\begin{equation}
	\label{MatrixT}
	{\mathbf{T}} =
	\left(
	\begin{array}{ccc}
    {\mathbf{t}_1}+{\mathbf{g}_1}&\cdots&{\mathbf{t}_m}+{\mathbf{g}_m}\\
	\end{array}
	\right)^T
	\end{equation}
	
	By combining Eq. (\ref{MatrixM}), Eq. (7), Eq. (\ref{MatrixA}), Eq. (\ref{MatrixT}), we upgrade single point transformation Eq. (\ref{TransformationFomulation}) to multiple points (model) transformation matrix formulation:\par 
	\begin{equation}
	\label{TransformationMatrix}
	E_{data}={\mathbf{R}_c}\cdot [{\mathbf{\Lambda}}\cdot{\mathbf{M}}+{\mathbf{T}} \cdot {\mathbf{C}}]+{\mathbf{T}_c} \otimes {\mathbf{1}}-{\hat{\mathbf{P}}} 
	\end{equation}
	where $\otimes$ is the kronecker product.\par 
	
 In SLAM problem formulation, the state vector is denoted as $X_i=[\mathbf{R}_{ci},\mathbf{T}_{ci},{\mathbf{\Lambda}}_i,\mathbf{T}_i]$ in $i$-th step.

	Robot to target measurement model: A typical SLAM observation model is a range and bearing model. In practice, there are several different measurements due to different sensors. Back-projection presentation is the most widely adopted observation model in RGBD and stereo SLAM \cite{song2018dynamic} \cite{song2018mis}. It is a modified version of closest points (ICP) taking advantage of regularized 2D depth observation, but in essence it is point to point pairing. For simplicity, we employ basic observation model, that is feature positions are directly observed by robot.\par 
	
	\subsection{Qualitative analysis of ED based SLAM formulation}
	\label{Proof_unobservability}
	
	We propose a counter-example to illustrate Eq. (\ref{TransformationMatrix}) is not observable. According to the definition \cite{yip1981solvability}, if there exists multiple optimal solutions $[\mathbf{R}_c,\mathbf{T}_c,{\mathbf{\Lambda}},{\mathbf{T}}]$ in one step transitional process (Eq. (\ref{TransformationMatrix})), it is at least partially not observable. Multiple optimal solutions lead to low-rank of the Fisher information matrix. The study of parameter observability examines whether the information provided by the available measurements is sufficient for estimating parameters without ambiguity \cite{bar2004estimation}. In other words, multiple solutions to the problem can be found attributing to insufficient information. Therefore, unobservability can be proved if multiple solutions to Eq. (\ref{TransformationMatrix}) are found, meaning global pose and non-rigid deformation formulation combined is not observable at the same time. \par 
	
	Here we show there are infinite solutions to Eq. (\ref{TransformationFomulation}). For the optimal solution [$\hat{\mathbf{R}}_c$ $\hat{\mathbf{T}}_c$ ${\hat{\mathbf{\Lambda}}}$ ${\mathbf{T}}$], we define an arbitrary rotation matrix $\mathbf{V_0}$. For a set of point cloud transformation (from ${\mathbf{P}}$ to ${\hat{\mathbf{P}}}$), Eq. (\ref{TransformationMatrix}) with the state vector [$\hat{\mathbf{R}}_c$ $\hat{\mathbf{T}}_c$ ${\hat{\mathbf{\Lambda}}}$ ${\mathbf{T}}$] can be reformulated into following form:
	
	{\setlength\abovedisplayskip{2pt} 
		\setlength\belowdisplayskip{2pt} 
		\begin{equation}
		\begin{aligned}
		{\hat{\mathbf{P}}}
		&=\hat{\mathbf{R}}_c [{\hat{\mathbf{\Lambda}}}{\mathbf{M}}+{\mathbf{T}}  {\mathbf{C}}]+\hat{\mathbf{T}}_c \otimes {\mathbf{1}} \\
		&=\hat{\mathbf{R}}_c \mathbf{V_0}  \mathbf{V_0}^T [{\hat{\mathbf{\Lambda}}}{\mathbf{M}}+{\mathbf{{\hat{T}}}}  {\mathbf{C}}]+\hat{\mathbf{T}}_c \otimes {\mathbf{1}}\\
		&=\hat{\mathbf{R}}_c \mathbf{V_0}  [\mathbf{V_0}^T  {\hat{\mathbf{\Lambda}}}{\mathbf{M}}+\mathbf{V_0}^T {\mathbf{{\hat{T}}}}  {\mathbf{C}}]+\hat{\mathbf{T}}_c \otimes {\mathbf{1}}\\
		\end{aligned}
		\end{equation}
				Therefore, there is a new solution [$\hat{\mathbf{R}}_c \mathbf{V_0}$, $\hat{\mathbf{T}}_c$, $\mathbf{V_0}^T  {\hat{\mathbf{\Lambda}}}$, $\mathbf{V_0}^T {\mathbf{T}}$]. Considering $\mathbf{V_0}$ is arbitrary, it's obvious that the incremental rigid rotation ${\mathbf{R}_c}$ can be offset by rotating the affine transformations matrix ${\mathbf{\Lambda}}$ in the opposite direction. For the \textbf{Rotation} constraint, ${\hat{\mathbf{\Lambda}}}$ is transformed by a rotation matrix which means $E_{rot}$ is unchanged. For the \textbf{Regularization} constraint:\par 
		{
			\begin{equation}
			E_{reg}=||\mathbf{V_0}^T[{\hat{\mathbf{\Lambda}}}{\mathbf{M}^{'}}+{\mathbf{T}} {\mathbf{C}} + {\mathbf{T}}]||_F^2,\\
			\end{equation}
			where ${\mathbf{M}^{'}}$ is similar to $\mathbf{M}$ with $\mathbf{v}_i$ ($i=1,...,n$) substituted by $\mathbf{g}_i$ ($i=1,...,m$). And $||\cdot||_F^2$ is the Frobenius norm. $E_{red}$ is a rotation of previous vector and the vector norm remains unchanged. In all, the new solutions [$\hat{\mathbf{R}}_c \mathbf{V_0}$, $\hat{\mathbf{T}}_c$, $\mathbf{V_0}^T  {\hat{\mathbf{\Lambda}}}$, $\mathbf{V_0}^T {\mathbf{T}}$] are also the optimal solutions to objective function Eq. (\ref{Eq:energyfunction}) in addition to [$\hat{\mathbf{R}}_c$ $\hat{\mathbf{T}}_c$ ${\hat{\mathbf{\Lambda}}}$ ${\mathbf{T}}$].\par 
			
			Similarly, for any arbitrary ${\Delta \mathbf{T}}$, we can find additional solutions satisfying Eq. (\ref{Eq:energyfunction}). Note that ${\Delta \mathbf{T}}\otimes {\mathbf{1}}={\mathbf{R}_c}{\Delta \mathbf{T}}\otimes {\mathbf{1}} {\mathbf{C}}$ due to the fact that the column sum of ${\mathbf{C}}$ is always to 1 (sum of weight). Thus we rewrite Eq. (\ref{TransformationMatrix}) to:
			\begin{equation}
			\begin{aligned}
			{\hat{\mathbf{P}}}
			&=\hat{\mathbf{R}}_c [{\hat{\mathbf{\Lambda}}}{\mathbf{M}}+{\mathbf{T}}  {\mathbf{C}}]+(\hat{\mathbf{T}}_c+{\Delta \mathbf{T}}-{\Delta \mathbf{T}}) \otimes {\mathbf{1}}\\
			&=\hat{\mathbf{R}}_c [{\hat{\mathbf{\Lambda}}}{\mathbf{M}}+{\mathbf{T}}  {\mathbf{C}}+{\Delta \mathbf{T}}\otimes ({\mathbf{1}} {\mathbf{C}})]+(\hat{\mathbf{T}}_c-{\Delta \mathbf{T}}) \otimes {\mathbf{1}}\\
			&=\hat{\mathbf{R}}_c [{\hat{\mathbf{\Lambda}}}{\mathbf{M}}+({\mathbf{T}}+{\Delta \mathbf{T}}\otimes {1}) {\mathbf{C}} ]+(\hat{\mathbf{T}}_c-{\Delta \mathbf{T}}) \otimes {1}\\
			\end{aligned}
			\end{equation}
			Accordingly, we have other solutions [$\hat{\mathbf{R}}_c$, $\hat{\mathbf{T}}_c-{\Delta \mathbf{T}}$, ${\hat{\mathbf{\Lambda}}}$, ${\mathbf{T}}+{\Delta \mathbf{T}}\otimes {1}$] to Eq. (\ref{Eq:energyfunction}). For the \textbf{Rotation} constraint, ${\hat{\mathbf{\Lambda}}}$ remains independent to ${\Delta \mathbf{T}}$. For the \textbf{Regularization} constraint:\par 
			
			\begin{equation}
			\begin{aligned}
			E_{reg}
			&=\sum_{j=1}^m\sum_{k\in{\mathbb{N}(j)}} \alpha_{jk}||\mathbf{A}_j(\mathbf{g}_k-\mathbf{g}_j)+\mathbf{g}_j+\mathbf{t}_j+{\Delta \mathbf{T}}-\\&(\mathbf{g}_k+\mathbf{t}_k+{\Delta \mathbf{T}})||^2\\
			&=\sum_{j=1}^m\sum_{k\in{\mathbb{N}(j)}} \alpha_{jk}||\mathbf{A}_j(\mathbf{g}_k-\mathbf{g}_j)+\mathbf{g}_j+\mathbf{t}_j-(\mathbf{g}_k+\mathbf{t}_k)||^2\\
			\end{aligned}
			\end{equation}
			Therefore, $E_{reg}$ remains the same for new solution [$\hat{\mathbf{R}}_c \mathbf{V_0}$, $\hat{\mathbf{T}}_c$, $\mathbf{V_0}^T  {\hat{\mathbf{\Lambda}}}$, $\mathbf{V_0}^T {\mathbf{T}}$].\par 
			
			\begin{remark}
			\label{remark:1}
			Prove is provided to show there are infinite number of optimal solutions to the energy function Eq. (\ref{Eq:energyfunction}). The global rotation matrix $\mathbf{R}_c$ or translation matrix $\mathbf{T}_c$ are entangled with ED parameters [${\mathbf{\Lambda}},{\mathbf{T}}$].
			\end{remark}
			
			\subsection{Proof of unobservability in ED based SLAM formulation}
			After a qualitative analysis, we provide observability analysis based on full Fisher information matrix analysis. Based on the discussion above, the unobservable lies in the $E_{data}$ defined in Eq. (\ref{TransformationMatrix}) with pairs $[\mathbf{R}_c,{\mathbf{\Lambda}}]$ and $[\mathbf{T}_c,{\mathbf{T}}]$, and is unrelated to Eq. (\ref{Erot}) and Eq. (\ref{RegulationConstraint}). Since global transformation parameters $\mathbf{R}_c$ and $\mathbf{T}_c$ are irrelevant to $E_{rot}$ and $E_{reg}$, the observability of these two terms are not affected. It's easy to prove that the partial Fisher information matrix with regard to $E_{rot}$ and $E_{reg}$, is full rank. Therefore, due to page limit, we only focus on the simplified case shown in Fig. \ref{fig:onesteponenode} with regard to Eq. (\ref{TransformationFomulation}) to analyze the observability. The conclusion of this node and one step robot movement can be generalized to multiple steps with a larger ED graph. Similarly, we prove that the low rank is located in information matrix with regard to $[\mathbf{R}_c,{\mathbf{\Lambda}}]$ and $[\mathbf{T}_c,{\mathbf{T}}]$. For simplification, we consider the residual of a single point $\mathbf{p}$ deformed by $m$ nodes to $\hat{\mathbf{p}}$:
			
				\begin{figure}[]
					\centering
					\includegraphics[width=0.48\textwidth]{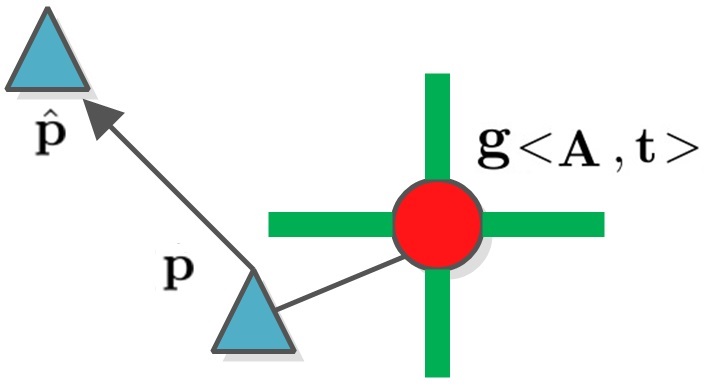}
					\caption{One step robot movement. Robot moves from $\mathbf{p}$ to $\hat{\mathbf{p}}$. The movement is a mixture of rigid transformation and deformation by ED node $\mathbf{g}$. The red line are the connecting edge from node $\mathbf{g}$ to other nodes.}
					\label{fig:onesteponenode}
				\end{figure} 
				
			\begin{equation}
			E_{data}^{'}={\mathbf{R}_c} [{\mathbf{\Lambda}}{\mathbf{M}}+{\mathbf{T}} {\mathbf{C}}]+{\mathbf{T}_c} \otimes {1} - \hat{\mathbf{p}} 
			\label{ObservabilityProve}
			\end{equation}
			
			where the state are  $[{\mathbf{\Lambda}}_m,{\mathbf{T}_m},{\mathbf{R}_c},{\mathbf{T}_c}]$. We vectorize the ${\mathbf{\Lambda}}_m$ and ${\mathbf{T}_m}$ and rewrite them into current form $[{\overline{\mathbf{\Lambda}}},{\overline{\mathbf{T}}},{\mathbf{R}_c},{\mathbf{T}_c]}$. Lie algebra is applied to optimize rotation matrix ${\mathbf{R}_c}$. For the convenience, we mark following variables:
			\begin{equation}
			\label{MatrixwidehatRc}
			{\underset{3\times 9m}{\mathbf{\widehat{R}_c}}} =
			\left[
			\begin{array}{ccc}
			{\mathbf{R}_c}&\cdots&{\mathbf{R}_c}\\
			\end{array}
			\right]
			\end{equation}
			\begin{equation}
			\label{MatrixwidetildeRc}
			{\underset{3\times 3m}{\mathbf{\widetilde{R}_c}}} =
			\left[
			\begin{array}{ccc}
			{\mathbf{R}_c}&\cdots&{\mathbf{R}_c}\\
			\end{array}
			\right]
			\end{equation}
			\begin{equation}
			{\underset{3\times 3m}{\mathbf{\widehat{C}}}} =
			\begin{bmatrix}
			{\mathbf{C}}&\cdots&{\mathbf{C}}\\
			\vdots&&\vdots\\
			{\mathbf{C}}&\cdots&{\mathbf{C}}\\
			\end{bmatrix}
			\end{equation}
			\begin{equation}
			\label{MatrixwidehatM}
			{\underset{3\times 9m}{\mathbf{\widehat{M}}}} =
			\begin{bmatrix}
			{\mathbf{M}}&\cdots&{\mathbf{M}}\\
			\vdots&&\vdots\\
			{\mathbf{M}}&\cdots&{\mathbf{M}}\\
			\end{bmatrix}
			\end{equation}
			
			\begin{equation}
			\underset{3\times 3}{\mathbf{S}}=skew({\mathbf{\Lambda}}\cdot{\mathbf{M}}+{\mathbf{T}} \cdot {\mathbf{C}})
			\end{equation}
				
			Where $skew(\cdot)$ is the skew symmetric operator. The Jacobian of Eq. (\ref{ObservabilityProve}) with regard to $[{\overline{\mathbf{\Lambda}}},{\overline{\mathbf{T}}},{\mathbf{R}_c},{\mathbf{T}_c]}$ is:
			\begin{equation}
			\label{J_prove}
			{\mathbf{J}} =
			\left(
			\begin{array}{cccc}
			\mathbf{\widehat{R}_c}\odot \mathbf{\widehat{M}}&\mathbf{\widetilde{R}_c}\odot \mathbf{\widehat{M}}&-{\mathbf{R}_c}{\mathbf{S}}&{\mathbf{I}}\\
			\end{array}
			\right)
			\end{equation}
			Where $\mathbf{I}$ is a 3 by 3 identity matrix. $\odot$ represents Hadamard product. Before estimating information matrix, we first mark the following matrix:
			\begin{equation}
			\label{MatrixA_1}
			{\underset{9m\times 9m}{\mathbf{A_1}}} =
			\begin{bmatrix}
			{\mathrm{1}}&\cdots&{\mathrm{1}}\\
			\vdots&&\vdots\\
			{\mathrm{1}}&\cdots&{\mathrm{1}}\\
			\end{bmatrix}
			\end{equation}
			\begin{equation}
			\label{MatrixA_2}
			{\underset{9m\times 3m}{\mathbf{A_2}}} =
			\begin{bmatrix}
			{\mathrm{1}}&\cdots&{\mathrm{1}}\\
			\vdots&&\vdots\\
			{\mathrm{1}}&\cdots&{\mathrm{1}}\\
			\end{bmatrix}
			\end{equation}
			\begin{equation}
			\label{MatrixA_3}
			{\underset{3m\times 3m}{\mathbf{A_3}}} =
			\begin{bmatrix}
			{\mathrm{1}}&\cdots&{\mathrm{1}}\\
			\vdots&&\vdots\\
			{\mathrm{1}}&\cdots&{\mathrm{1}}\\
			\end{bmatrix}
			\end{equation}
			\begin{equation}
			\label{MatrixA_4}
			{\underset{9m\times 3}{\mathbf{A_4}}} =
			\begin{bmatrix}
			{\mathrm{1}}\\
			\vdots\\
			{\mathrm{1}}\\
			\end{bmatrix}
			\end{equation}
			\begin{equation}
			\label{MatrixA_5}
			{\underset{3m\times 3}{\mathbf{A_5}}} =
			\begin{bmatrix}
			{\mathrm{1}}\\
			\vdots\\
			{\mathrm{1}}\\
			\end{bmatrix}
			\end{equation}
			Based on all the definitions, the Hessian matrix ${\boldsymbol{\mathcal{H}}_{ed}}$ can be presented in the following form:
			\begin{equation}
			\arraycolsep=2.5pt\def\arraystretch{1.8}
			\begin{aligned}
			&{\boldsymbol{\mathcal{H}}_{ed}} =
			\begin{bmatrix}
			{\mathbf{H}_1}\\
			{\mathbf{H}_2}\\
			{\mathbf{H}_3}\\
			{\mathbf{H}_4}\\
			\end{bmatrix}
			\stackrel{\mathrm{def}}{=}
			\\
			&\begin{bmatrix}
			\begin{matrix}\mathbf{A_1}\odot\\ (\mathbf{\widehat{M}}^T\mathbf{\widehat{M}})\end{matrix}&\begin{matrix}\mathbf{A_2}\odot\\ (\mathbf{\widehat{M}}^T\mathbf{\widehat{C}})\end{matrix}&\begin{matrix}-\mathbf{\widehat{M}}^T\odot\\ (\mathbf{A_4}\mathbf{S})\end{matrix}&\mathbf{\widehat{R}_c}^T\odot \mathbf{\widehat{M}}^T\\
			\begin{matrix}\mathbf{A_2}\odot\\ (\mathbf{\widehat{C}}^T\mathbf{\widehat{M}})\end{matrix}&\begin{matrix}\mathbf{A_3}\odot\\ (\mathbf{\widehat{C}}^T\mathbf{\widehat{C}})\end{matrix}&\begin{matrix}-\mathbf{A_5}\odot\\ (\mathbf{\widehat{C}}^T\mathbf{S})\end{matrix}&\mathbf{\widetilde{R}_c}^T\odot \mathbf{\widehat{M}}^T\\
			\begin{matrix}-(\mathbf{S}^T\mathbf{A_4}^T)\\ \odot \mathbf{\widehat{M}}\end{matrix}&\begin{matrix}-(\mathbf{S}^T\mathbf{A_5}^T)\\ \odot \mathbf{\widehat{C}}\end{matrix}&\mathbf{S}^T\mathbf{S}&-\mathbf{S}^T\mathbf{R}_c^T\\
			\mathbf{\widehat{R}_c}\odot \mathbf{\widehat{M}}&\mathbf{\widetilde{R}_c}\odot \mathbf{\widehat{M}}&-{\mathbf{R}_c}{\mathbf{S}}&{\mathrm{I}}\\
			\end{bmatrix}
			\end{aligned}
			\end{equation}
			For the sub matrix $\mathbf{H}_1$ and $\mathbf{H}_2$ within the Hessian matrix ${\boldsymbol{\mathcal{H}}_{ed}}$, we split them into the group of every 3 lines. For example, $\mathbf{H}_1(i)$ is the group $i$ ranging from line $3*(i-1)+1$ to $3i$. By analyzing Hessian matrix ${\boldsymbol{\mathcal{H}}_{ed}}$, we discover the following law:
			 
			\begin{equation}
			\mathbf{H}_1(i)=\dot{\mathbf{M}}\odot[-(\mathbf{S}^T)^{-1}\mathbf{H}_3]
			\label{Law1}
			\end{equation}
			\begin{equation}
			\mathbf{H}_2(i)=\dot{\mathbf{C}}\odot[\mathbf{R}_c^T\mathbf{H}_4]
			\label{Law2}
			\end{equation}
			where $\dot{\mathbf{M}}$ and $\dot{\mathbf{C}}$ are defined in the following form:
			
			\begin{equation}
			\label{MatrixdotM}
			\underset{3\times 18m}{\dot{\mathbf{M}}} =
			\begin{bmatrix}
			\mathbf{\widehat{M}}&\mathbf{\widehat{M}}&\mathbf{\widehat{M}}\\
			\end{bmatrix}
			\end{equation}
			
			\begin{equation}
			\label{MatrixdotC}
			{\underset{3\times 18m}{\mathbf{\dot{C}}}} =
			\begin{bmatrix}
			{\mathbf{\widehat{C}}}&\cdots&{\mathbf{\widehat{C}}}\\
			\end{bmatrix}
			\end{equation}
			
			Obviously, this one point transformation scenario can be extended to multiple points. Eq. (\ref{Law1}) and Eq. (\ref{Law2}) indicate that the global rotation ${\mathbf{R}_c}$ matrix and translation vector ${\mathbf{T}_c}$ can be embedded into local affine deformation matrix  ${\mathbf{\Lambda}}]$ and ${\mathbf{T}}$ respectively. This conclusion also validates the qualitative conclusions (Remark \ref{remark:1}) we draw in Section \ref{Proof_unobservability}.

			\section{Priori based SLAM formulation}
			\label{section:priorbasedSLAM}
			Section \ref{section:observability} shows the inner-connection between the rigid relative transformation and the non-rigid deformation formulations. The two pairs, $(\mathbf{R}_c,{\mathbf{\Lambda}})$ and $(\mathbf{T}_c,\mathbf{T})$, are intertwined. Thus, both global rotation and translation cannot be uniquely determined in conventional ED formulation on condition that no new information is provided. There are infinite number of solutions to the robot poses if robot to feature observations is the only source of input. \textbf{This conclusion can be generalized to other deformation formulation like FEM or structure-from-template because the degree-of-freedom of deformation enables rigid model motion just with deliberately adjusted movement of model vertices.} With regard to this, the goal is to propose a prior to separate and determine the relative rigid transformation from the deforming non-rigid tissue. Noteworthily, rigid SLAM algorithm with thresholds based feature classification strategy \cite{mahmoud2016orbslam,mahmoud2017slam,turan2017non} also comes with prior, assuming the static features are known and can be verified with thresholds. In this work, however, we still assumes all non-rigid surfaces are deforming. Experiments in Section \ref{section:resultsanddisucssion} demonstrate that information matrix is full rank and estimated parameters are unique with the proposed priori.\par 
			
		In the field of non rigid structure from motion, features are granted more freedom under the base shape constraints \cite{agudo2017force} \cite{agudo2018robust}. Instead of one single static position in pose estimation, features are formulated with 3D locations in each frame. To prevent the irregular movement of the 3D shapes, base shapes \cite{garg2013dense}, base trajectories \cite{valmadre2012general} or base shape-trajectory \cite{simon2014separable} strategies are introduced to limit the degree of freedom of the soft shapes. They assume that the movement is a mixutre of predefined bases, although these predefined bases are also unknown for the observation. After enforcing the bases, the freedom of the deformation is constrained and the rigidity of deformation can be controlled by the number of all the bases.\par 
		Taking advantage of this, we propose that deformation of the feature can be approximated by base historical shapes and the residuals of base shapes approximation is the rigid robot movement. Theoretically, if provided with infinite number of base shapes, deformation of features as well as robot can be accurately estimated. In practice, comparing with traditional rigid SLAM or ED based SLAM, limited number of base shapes can still generate good robot pose due to observability preserved. This is especially true in complex periodic deformation scenario where deformation is caused by breathing and heartbeat; current deformed shape can be inferred from previous shapes. \par
			
			Based on the proposed prior, a new feature motion formulation is introduced in the conventional back-end rigid SLAM formulation. In our study, the primary feature motion measurement is based on the idea that current structure $\mathbf{f}^{n+1}$ can be linearly fitted by its historical shapes ${\mathbf{f}^n}\ ...\  {\mathbf{f}^{n-t}}$ where $t$ is the processing window. A coefficient vector ${\mathbf{c}}=[{\delta_1} \ ,...,\ {\delta_t}]$ is introduced to describe the relations of these feature movements. Matrix ${\mathbf{B}}$ ($3N\times F$) is the combination of all valid features. $N$ is the number of features and $F$ is the number of steps. Note that some elements in ${\mathbf{B}}$ is invalid because the viewing angle of robot makes it unable to observe all features at all steps.\par 
			
			\begin{equation}
			\label{MatrixB}
			{\mathbf{B}} =
			\begin{bmatrix}
			{\mathbf{f}}_1^1&{\mathbf{f}_1^2}&\cdots&\cdots&{\mathbf{f}_1^F}\\
			\vdots&\vdots&\vdots&\vdots&\vdots\\
			{\mathbf{f}}_N^1&{\mathbf{f}_N^2}&\cdots&\cdots&{\mathbf{f}_N^F}\\
			\end{bmatrix}
			\end{equation}
			
			The term `validity' of feature ${\mathbf{f}_i^j}$ in ${\mathbf{B}}$ refers to (1) feature $i$ is observed by robot in step $j$ and (2) feature $i$ is observed in the period window $t$; in other words $[{\mathbf{f}_{i}^j}...{\mathbf{f+t}_{i}^j}]$ are all observed by robot. The validity ensure building correlations in a consecutive movement of feature. \par

			\begin{table}[!h]
				\caption{Pose and feature errors in Monte Carlo simulations.}
				\label{Table:Monte_Carlo_Simulation}
				\begin{center}
					\begin{tabular}{|p{2.8cm}|p{1.4cm}|p{1.4cm}|p{1.4cm}|}
						\hline
						& Deformable SLAM (m) & Least Square (m) & ED node based VO (m) \\
						\hline
						\multicolumn{4}{|c|}{Simulation 1}\\
						\hline
						Robot Position X(m) &0.942&2.538 & 8.743\\
						\hline
						Robot Position Y(m)&0.526&1.012 & 3.197\\
						\hline
						Robot Heading (Rad)&0.005 &0.009 & 0.014\\
						\hline
						\multicolumn{4}{|c|}{Simulation 2}\\
						\hline
						Robot Position X(m) &0.119 &0.277&2.098 \\
						\hline
						Robot Position Y(m)&0.138  &0.498& 3.38 \\
						\hline
						Robot Heading (Rad)&0.002 &0.002& 0.009\\
						\hline
					\end{tabular}
				\end{center}
				\label{Table:monte_carlo_results}
			\end{table}
			\begin{table}[!h]
				\caption{Pose and feature errors in of heart, stomach and lung.}
				\label{Table:Monte_Carlo_Simulation}
				\begin{center}
					\begin{tabular}{|p{2.8cm}|p{1.4cm}|p{1.4cm}|p{1.4cm}|}
						\hline
						& Deformable SLAM & Rigid SLAM & ED node based VO \\
						\hline
						\multicolumn{4}{|c|}{Heart scenario}\\
						\hline
						Robot Position X(unit) &0.149& 2.006 & 8.743\\
						\hline
						Robot Position Y(unit)&0.085&0.951 & 3.197\\
						\hline
						Robot Heading (Rad)&0.001 &0.001 & 0.010\\
						\hline
						\multicolumn{4}{|c|}{Stomach scenario}\\
						\hline
						Robot Position X(unit) &2.263 &7.004&2.098\\
						\hline
						Robot Position Y(unit)&2.566  &6.894& 3.380 \\
						\hline
						Robot Heading (Rad)&0.006\ &0.008& 0.009 \\
						\hline
						\multicolumn{4}{|c|}{Lung scenario}\\
						\hline
						Robot Position X(unit) &2.009 &7.596&2.098\\
						\hline
						Robot Position Y(unit)&0.706  &3.304& 3.380 \\
						\hline
						Robot Heading (Rad)&0.002 &0.003& 0.009 \\
						\hline
					\end{tabular}
				\end{center}
			\end{table}

			The proposed formulation is based on conventional back-end rigid SLAM. We first introduce rigid SLAM here. In 3D scenario where one robot freely moves with $N$ static features, the state to be estimated is denoted as:\par  
			\begin{equation}
			\mathbf{X}=[\mathbf{R}\  \mathbf{p}\ \mathbf{f}_1\ \cdots\ \mathbf{f}_N],
			\end{equation}
			where $\mathbf{R}\in{\mathbb{SO}(3)}$ is the robot orientation, $\mathbf{p}\in{\mathbb{R}^3}$ is the robot position, $\mathbf{f}_i\in{\mathbb{R}^3}$ is the $i$th feature (ranging from $1$ to $N$). The general rigid robot motion model from step $n$ to $n+1$ without noise is described as:
			
			\begin{equation}
			\label{RobotStateTransition}
			\begin{aligned}
			&\mathbf{p}^{n+1}=\mathbf{p}^n+\mathbf{R}^n \mathbf{v}^n\\
			&\mathbf{R}^{n+1}=\mathbf{R}^n  \mathbf{\omega}^n\\
			\end{aligned}
			\end{equation}
			where $\mathbf{v}^n$ is the linear translation of one step movement. $\mathbf{\omega}^n \in{\mathbb{SO}(3)}$ is the rotation matrix describing orientation variation.  \par

			The rigid SLAM formation is modified by applying the time series method. When depicting feature motion, we are bereft of analogy of conventional feature movement, so our implementation is to build relationship of a given feature in consecutive movement. The formulation manoeuvres to constrain feature motion to a mixture of historical movement. The constraint of feature motion model is expressed by building linear relations within a window of feature locations. The main advantage of linearly modelling the feature locations over historic base shape modelling is that it can initialize feature locations with rigid assumption (using conventional visual odometry) and avoids base feature recognition. In non-rigid structure from motion, base shapes are essential to describe deformation. Moreover, base shapes require different window sizes for modelling which poses heavy computational burden. The proposed linear constraint, however, is flexible and straightforward to complex mixed deformation. In addition to the robot motion model Eq. (\ref{RobotStateTransition}), the proposed linear feature motion is modelled as:\par 
			\begin{equation}
			\label{Eq:FeatureConnection}
			\begin{aligned}
			\mathbf{f}_i^{n+1}={{\delta_1}\cdot\mathbf{f}_{i}^{n}}+{{\delta_2}\cdot\mathbf{f}_{i}^{n-1}}+\cdots+{{\delta_t}\cdot\mathbf{f}_{i}^{n-t}}
			\end{aligned}
			\end{equation}

			\subsection{Prediction modelling}
			We modify the conventional state to $[\mathbf{R}^1,\mathbf{p}^1,...,\mathbf{R}^n,\mathbf{p}^n,\mathbf{B},\mathbf{c}]$. 
			
			\begin{equation}	\argmin\limits_{\mathbf{R}^1,\mathbf{p}^1,...,\mathbf{R}^n,\mathbf{p}^n,\mathbf{B},\mathbf{c}} E_{obs}+E_{f}+E_{ini}
			\label{energyfunction}
			\end{equation}
			
			Eq. (\ref{energyfunction}) is the energy function for a visual deformable SLAM. $E_{obs}$ is the sum error of robot to feature observations:
			
			\begin{equation}
			E_{obs}=\sum_{i=1}^N\sum_{j=1}^F[{\mathcal{F}}(\mathbf{R}^j,\mathbf{p}^j,\mathbf{f}_{i}^{j})-\mathbf{m}_i^j]^2, 
			\end{equation}
			where $\mathbf{m}_i^j$ is the observation from robot to location of feature $i$ in step $j$. ${\mathcal{F}}(\cdot)$ encodes the estimated observation from robot pose to feature position.\par 
			
			$E_f$ denotes the error between current feature and its estimation from historical locations following Eq. (\ref{Eq:FeatureConnection}):
			\begin{equation}
			\begin{aligned}
			E_f=&\sum_{i=1}^N\sum_{j=1}^t({\mathbf{f}_i^{j+1}}-{{\delta_1}\cdot\mathbf{f}_{i}^{j}}-{{\delta_2}\cdot\mathbf{f}_{i}^{j-1}}\\
			&-\cdots-{{\delta_t}\cdot\mathbf{f}_{i}^{j-t}})^2\\
			\end{aligned}
			\end{equation}
			
			\begin{equation}
			E_{ini}=\sum_{i=1}^t[\mathbf{p}^i-\mathbf{p}^0]^2+\sum_{i=1}^t[\mathbf{R}^i \ominus \mathbf{R}^0]^2 
			\end{equation}
			
			$E_{ini}$ is to ensure the initial robot pose keeps static in the period size $t$. The notation $\ominus$ is called inverse retraction
			in differentiable geometry \cite{absil2007trust} and it is designed as a smooth
			mapping such that $\mathbf{R}=\mathbb{R} \ominus \mathbf{0}$. Similar to conventional rigid SLAM problem where the first pose need to be fixed \cite{zhang2017convergence}, in our formulation the first period of poses should be fixed likewise.\par 
			
			\begin{figure}[]
				\centering
				\includegraphics[width=0.48\textwidth]{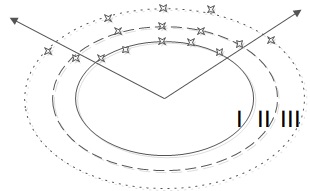}
				\caption{A typical feature deforming example. The ellipse deforms periodically depicted in 'I, II and III'. The region within arrows are the visible region. The leftmost feature is not observed in phase 'II' and 'III'. The rightmost feature is not observed in phase 'I'.}
				\label{fig:featurenotseen}
			\end{figure}  
			
			Due to field of view of camera, some of the
			features may not be seen when the environment deforms. Fig. \ref{fig:featurenotseen} shows one example of features not seen in some steps. Our approach is capable of processing this situation. If one feature is not fully observed any sliding window like the example shows, we will ignore this feature.
			
			\subsection{Observability analysis}
			\label{observability_test_deformable}
			In this section, we examine the parameter observability properties the proposed deformable SLAM formulation, which, for the time being, is considered as a parameter estimation problem. We will prove that the coefficient matrix ${\mathbf{c}}$ is not observable but the robot pose as well as feature motions are observable. This is a very satisfying result because coefficient matrix ${\mathbf{c}}$ is only an auxiliary variable and is not physically explainable in real scenario. Robot pose as well as feature motions, however, are physical processes and needs to be accurately estimated.\par 
			
			We adduce examples to prove coefficient matrix ${\mathbf{c}}$ is not observable. Taking into account the flexibility of presenting multiple period motions in Eq. (\ref{Eq:FeatureConnection}), it will inevitably result in multiple solutions of feature motion combination. When features are static, current shape of the environment will be passed to next formulation which means all $\mathbf{c}=[1, 0 \ ,...,\ 0]$. When there's only one periodic movement, the shape of the environment will be the same shape in history $\mathbf{c}=[0, \ ,...0,1,0,...,\ 0]$. In more general scenario, multiple periodic movement will lead to full $\mathbf{c}$. 
			We would like to emphasize that: The positions of features are not solvable. A simple example is when period is 2 but window is 4, this will be presented by $\mathbf{c}=[0,1,0,0]$ or $\mathbf{c}=[0,0.5,0,0.5]$. \par

			\begin{figure}[]
				\centering
				\includegraphics[width=0.5\textwidth]{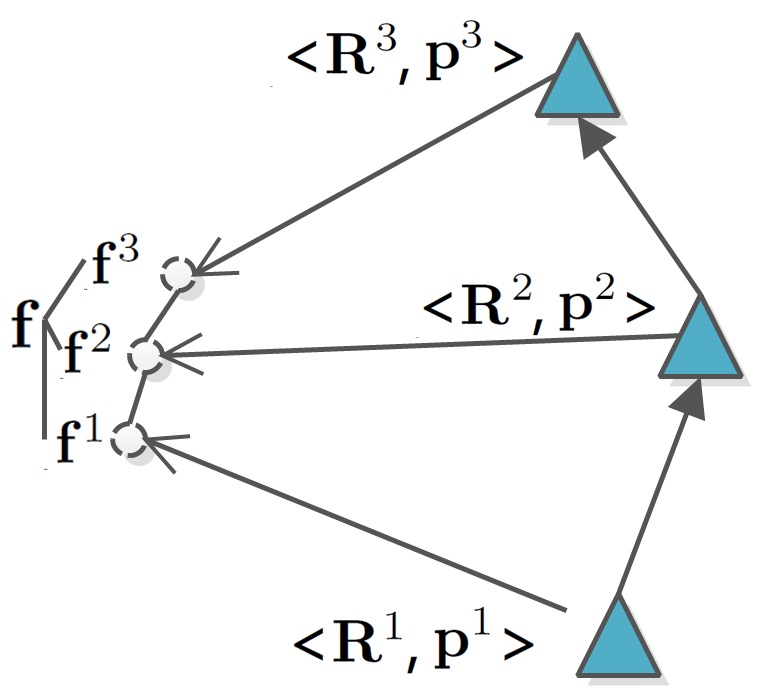}
				\caption{A simple example of 2 steps robot movement. Different from SLAM in rigid scenario, the feature $\mathbf{f}$ deforms in the space in position $\mathbf{f}^1$, $\mathbf{f}^2$ and $\mathbf{f}^3$.}
				\label{fig:timeseries_example}
			\end{figure}

			In addition to qualitative analysis of observability, we gain a better understanding of the formulation by proving with definition of observability. The study of parameter observability examines whether the information provided by the available measurements
			is sufficient for estimating the parameters without ambiguity; when parameter observability holds, the Fisher information matrix (FIM) is full rank and invertible \cite{bar2004estimation}. From the stated example we have obtained the idea that the unobservable part lies in mismatch of real period and predefined window size. Therefore, we first prove this with simple scenario and extended to our conclusion. Consider the scenario shown in Fig. \ref{fig:timeseries_example}, one robot moves in three steps with orientation $\mathbf{R}^1$, $\mathbf{R}^2$, $\mathbf{R}^3$ and position $\mathbf{p}^1$, $\mathbf{p}^2$, $\mathbf{p}^3$. And it always observe one deforming feature with position $\mathbf{f}^1$, $\mathbf{f}^2$, $\mathbf{f}^3$. The observation is $\mathbf{z}_1$, $\mathbf{z}_2$ and $\mathbf{z}_3$. Window size $t$ is set to 2. The objective function should be:
			
			\begin{equation}
			\label{Eq:F_minimalExample}
			{\mathrm{F}_{obj}} =
			\begin{bmatrix}
			\mathbf{R}^1\cdot(\mathbf{f}^1-\mathbf{p}^1)-\mathbf{z}_1\\
			\mathbf{R}^2\cdot(\mathbf{f}^2-\mathbf{p}^2)-\mathbf{z}_2\\
			\mathbf{R}^3\cdot(\mathbf{f}^3-\mathbf{p}^3)-\mathbf{z}_3\\
			{{\mathbf{f}^3}}-{{\delta_1}\cdot{\mathbf{f}^1}}-{{\delta_2}\cdot{\mathbf{f}^2}}\\
			\mathbf{R}^1 \ominus \mathbf{I}_3\\
			\mathbf{R}^2 \ominus \mathbf{I}_3\\
			\mathbf{p}^1\\
			\mathbf{p}^2\\
			\end{bmatrix}
			\end{equation}
			where $\ominus$ defines the distance of in the space of $SO(3)$ meaning the first two orientation $\mathbf{R}^1$ and $\mathbf{R}^2$ is fixed (close to $3\time3$ identity matrix $\mathrm{1}_3$). The corresponding Jacobian of the toy model Eq. (\ref{Eq:F_minimalExample}) is shown in Eq. (\ref{J_minimalExample}). $S(\cdot)$ is the skew symmetric formulation. Therefore, the corresponding FIM matrix is Eq. (\ref{JtJ_minimalExample}). With regard to this scenario, after Gaussian elimination, Matrix $H$ is full rank if the feature is moving ($\mathbf{f}^1$, $\mathbf{f}^2$ and $\mathbf{f}^3$ are not equivalent). However, considering the last $5 \time 5$ of matrix $H$, when feature is stable, all feature poses $\mathbf{f}^1$, $\mathbf{f}^2$ and $\mathbf{f}^3$ are equivalent and coefficients ${\delta_1}$ and ${\delta_2}$ are the same. In this case, $H$ loses one rank thanks to the last two lines of matrix $H$. Thus, the only contribution to low rank lies in the last two lines of matrix $H$ corresponding to variable coefficients ${\delta_1}$ and ${\delta_2}$ and is irrelevant to the number of features and number of steps.  On the basis of these analysis we concluded that in general scenario, the low rank of Hessian is contributed by coefficients in the case of all features are stable.\par

			\setlength\arraycolsep{4pt}
			\begin{figure*}[t]
				\normalsize
				\setcounter{mytempeqncnt}{\value{equation}}
				\setcounter{equation}{40}
			\begin{equation}
			\arraycolsep=2.4pt\def\arraystretch{2.5}
			\label{J_minimalExample}
			\boldsymbol{\mathcal{J}} =
			\begin{bmatrix}
			-\mathbf{R}^1\cdot S(\mathbf{f}^1-\mathbf{p}^1)&-\mathbf{R}^1&&&&&\mathbf{R}^1&&&\\
			&&-\mathbf{R}^2\cdot S(\mathbf{f}^2-\mathbf{p}^2)&-\mathbf{R}^2&&&&\mathbf{R}^2&&\\
			&&&&-\mathbf{R}^3\cdot S(\mathbf{f}^3-\mathbf{\mathbf{p}}^3)&-\mathbf{R}^3&&&\mathbf{R}^3&\\
			&&&&&-{\delta_1}\cdot \mathbf{I}_3&-{\delta_2}\cdot \mathbf{I}_3& \mathbf{I}_3&-\mathbf{f}^1&-\mathbf{f}^2\\
			-\mathbf{R}^1&\mathbf{I}_3&&&&&&&&\\
			&&-\mathbf{R}^2&\mathbf{I}_3&&&&&&\\
			\end{bmatrix}
			\end{equation}
				\setcounter{equation}{\value{mytempeqncnt}}
				\vspace*{4pt}
			\end{figure*}

	\begin{figure*}
		\normalsize
		\setcounter{mytempeqncnt}{\value{equation}}
		\setcounter{equation}{41}
		\begin{equation}
		\setcounter{MaxMatrixCols}{20}
		\arraycolsep=0.01pt\def\arraystretch{2.5}
		\label{JtJ_minimalExample}
		\begin{aligned}
		&{\boldsymbol{\mathcal{H}}} =\\
		&\begin{bmatrix}
		2  \mathbf{I}_3&\begin{matrix}S(\mathbf{f}^1\\-\mathbf{p}^1)\end{matrix}&&&&&-S(\mathbf{f}^1-\mathbf{p}^1)&&&&\\
		2  \mathbf{I}_3&\begin{matrix}S(\mathbf{f}^1\\-\mathbf{p}^1)\end{matrix}&&&&&-S(\mathbf{f}^1-\mathbf{p}^1)&&&&\\
		S(\mathbf{f}^1-\mathbf{p}^1)&2  \mathbf{I}_3&&&&&-\mathbf{I}_3&&&&\\
		&&2  \mathbf{I}_3&S(\mathbf{f}^2-\mathbf{p}^2)&&&&-S(\mathbf{f}^2-\mathbf{p}^2)&&&\\
		&&S(\mathbf{f}^2-\mathbf{p}^2)&2  \mathbf{I}_3&&&&-\mathbf{I}_3&&&\\
		&&&&2  \mathbf{I}_3&S(\mathbf{f}^3-\mathbf{p}^3)&&&-S(\mathbf{f}^3-\mathbf{p}^3)&&\\
		&&&&S(\mathbf{f}^3-\mathbf{p}^3)&2  \mathbf{I}_3&&&-\mathbf{I}_3&&\\
		-S(\mathbf{f}^1-\mathbf{p}^1)&-\mathbf{I}_3&&&&&{\delta_1}^2  \mathbf{I}_3&{\delta_1} {\delta_2}  \mathbf{I}_3&-{\delta_1}  \mathbf{I}_3&{\delta_1}  \mathbf{f}^1&{\delta_1}  \mathbf{f}^2\\
		&&-S(\mathbf{f}^2-\mathbf{p}^2)&-\mathbf{I}_3&&&{\delta_1} {\delta_2}  \mathbf{I}_3&{\delta_2}^2  \mathbf{I}_3&-{\delta_2}  \mathbf{I}_3&{\delta_2}  \mathbf{f}^1&{\delta_2}  \mathbf{f}^2\\
		&&&&-S(\mathbf{f}^3-\mathbf{p}^3)&-\mathbf{I}_3&-{\delta_1}  \mathbf{I}_3&-{\delta_2}  \mathbf{I}_3& \mathbf{I}_3&-\mathbf{f}^1&-\mathbf{f}^2\\
		&&&&&&{\delta_1}  {\mathbf{f}^1}^T&{\delta_2}  {\mathbf{f}^1}^T&-{\mathbf{f}^1}^T\mathbf{I}_3&{\mathbf{f}^1}^T*\mathbf{f}^1&{\mathbf{f}^1}^T*\mathbf{f}^2\\
		&&&&&&{\delta_1}  {\mathbf{f}^2}^T&{\delta_2}  {\mathbf{f}^2}^T&-{\mathbf{f}^2}^T\mathbf{I}_3&{\mathbf{f}^2}^T*\mathbf{f}^1&{\mathbf{f}^2}^T*\mathbf{f}^2\\
		\end{bmatrix}
		\end{aligned}
		\end{equation}
		\setcounter{equation}{\value{mytempeqncnt}}
		\vspace*{4pt}
	\end{figure*} 		
			
%
%
			
			\begin{figure*}[!h]
				\centering
				\subfloat{	
					\begin{minipage}[]{0.33\textwidth}	
						\centering
						\includegraphics[width=1\linewidth]{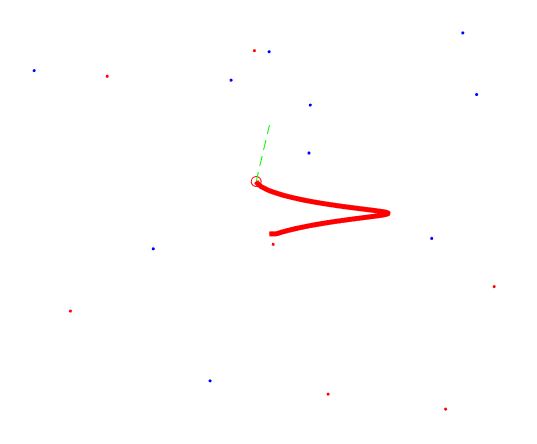}			
					\end{minipage}	
				}\/
				\subfloat{	
					\begin{minipage}[]{0.33\textwidth}	
						\centering
						\includegraphics[width=1\linewidth]{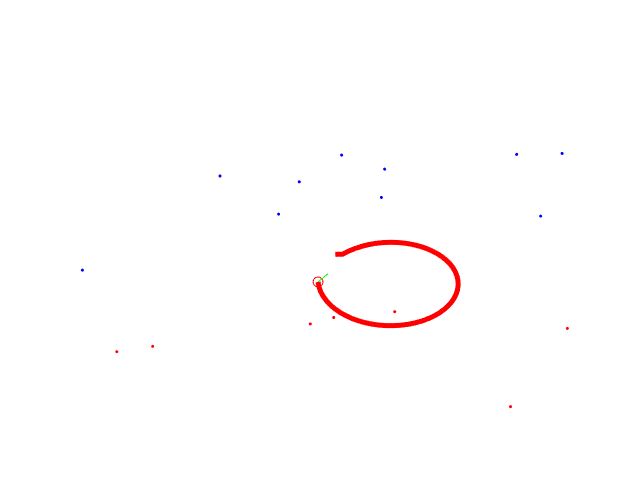}			
					\end{minipage}	
				}
				\caption{The two figures is an example of Monte Carlo simulation. Display area is illustrated from different directions for visualization.}
				\label{fig:trajectory_2models}
			\end{figure*}	
			\begin{figure*}[!h]
				\centering
				\subfloat[A deformable heart]{
					\begin{minipage}[]{0.3\textwidth}
						\centering
						\includegraphics[width=1\linewidth]{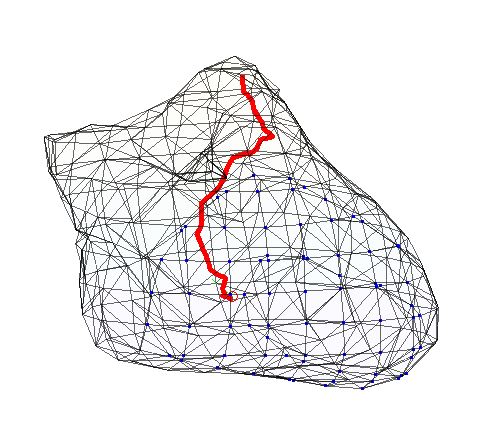}
					\end{minipage}
				}\/
				\subfloat[Trajectory in a deformable stomach]{
					\begin{minipage}[]{0.3\textwidth}
						\centering
						\includegraphics[width=1\linewidth]{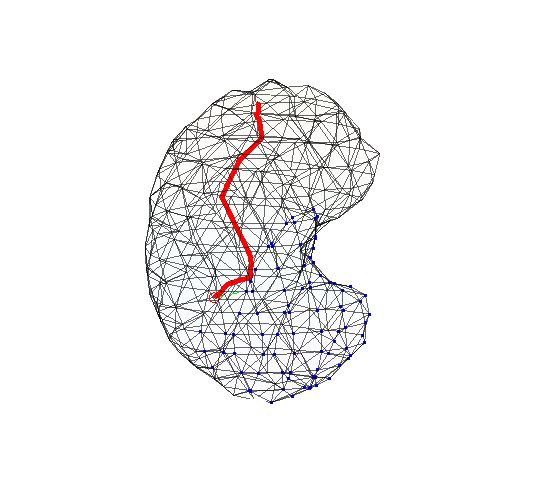}
					\end{minipage}
				}\/
				\subfloat[Trajectory in a deformable liver]{
					\begin{minipage}[]{0.3\textwidth}
						\centering
						\includegraphics[width=1\linewidth]{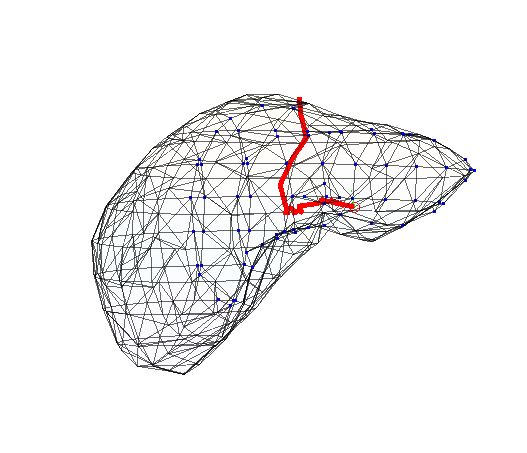}
					\end{minipage}
				}
				\caption{(a), (b) and (c) shows the robot moves randomly inside a deformable organ (Heart, stomach and liver). Red curves are the trajectories. Blue dots are the positions of the features and the attached quiver is the corresponding moving direction of each feature. Quiver only shows one step. Please refer to our video for the whole process.}
				\label{fig:trajectory_3models}
			\end{figure*}
			
			\begin{figure*}[]
				\centering
				\includegraphics[width=1\textwidth]{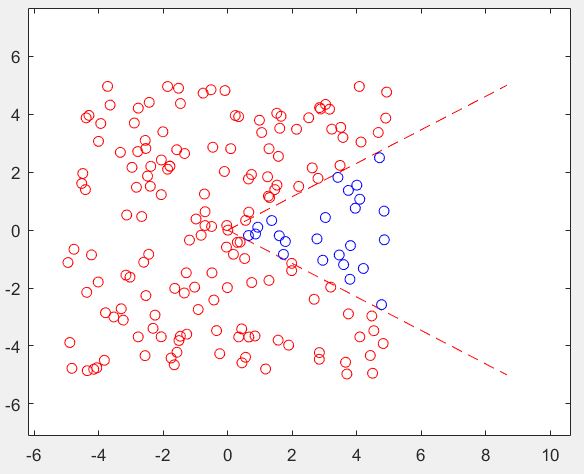}
				\caption{Estimation errors of rigid SLAM and the proposed time-series SLAM. Row 1 to 3 are the tests on scenarios of heart, liver and left lung. Column 1 to 3 are the RMSE of robot position X, robot position Y and robot heading.}
				\label{fig:RMSE_MonteCarlo}
			\end{figure*}  	
			
			\begin{figure*}[]
				\centering
				\includegraphics[width=1\textwidth]{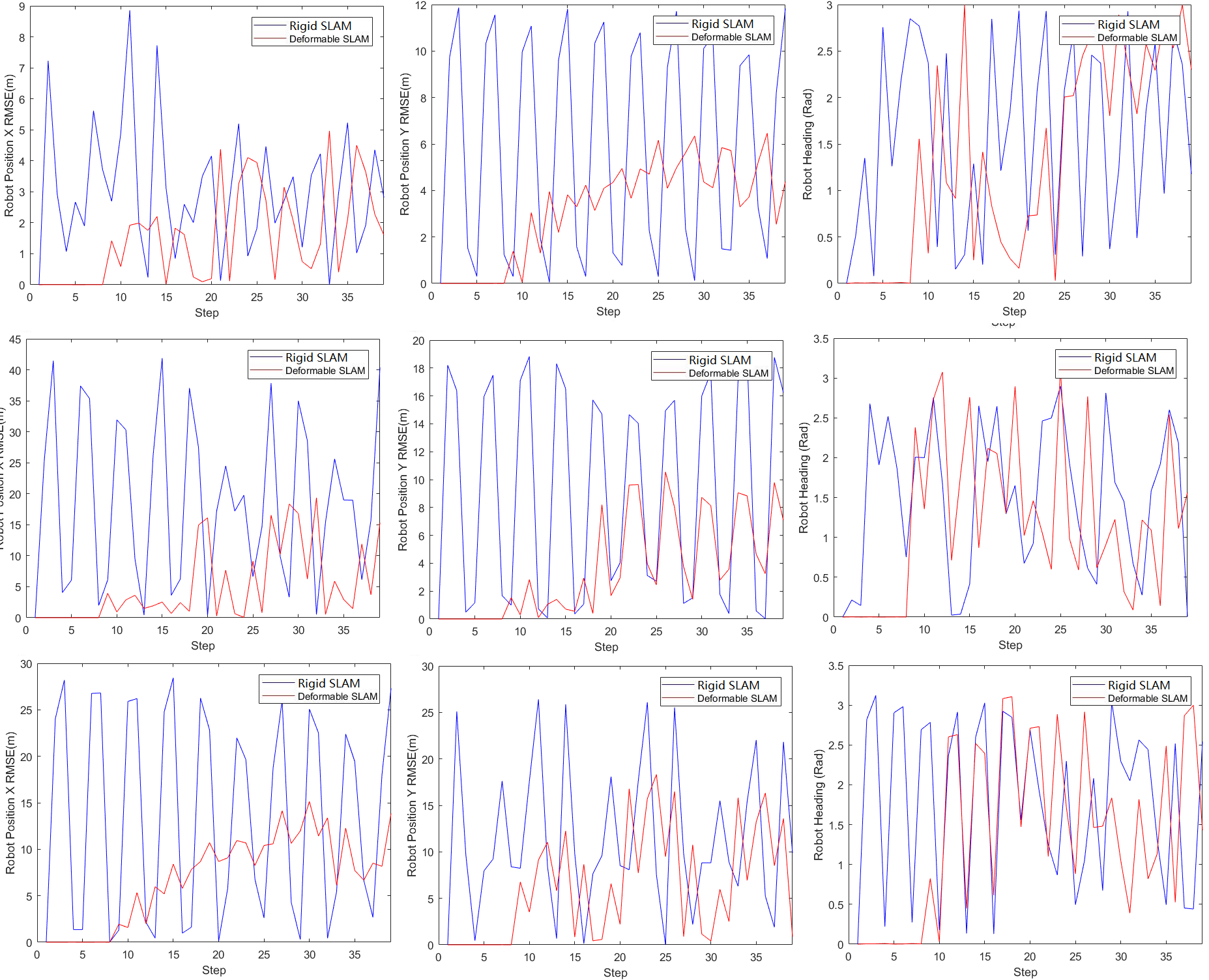}
				\caption{Estimation errors of rigid SLAM and the proposed time-series SLAM. Row 1 to 3 are the tests on scenarios of heart, liver and left lung. Column 1 to 3 are the RMSE of robot position X, robot position Y and robot heading.}
				\label{fig:RMSE_3models}
			\end{figure*}  
			
			\section{Results and discussion}
			\label{section:resultsanddisucssion}
			\subsection{Monte Carlo simulations}
			In order to validate the proposed priori based approach as well as prove the unobservable robot pose in deformable scenario, we conduct a series of Monte Carlo simulations under various conditions like different period of deformation, different movement of robots and different visibility of robot to feature observations. Fig. \ref{fig:trajectory_2models} is the typical 3D robot movements with 20 deforming features and 60 steps. The observation is defined as feature position in robot coordinate which is very common scenario of either stereoscope, lidar, RGBD and stereo camera sensors. We adopt a deformation generator to simulate mixed kinematic deformations (different period and amplitude) of the features. The simulation size ranges $500 \times 500$ (mm). Robot moves in a predefined trajectory with 20 features deforming in a randomly mixed periodic way imitating soft-tissue movement. The viewing angle is also randomly chosen ranging from $30^\circ$ to $90^\circ$. Noises are imposed on robot to feature observation ranging from 1 to 5 mm. In this test we just focus on adverse robot-feature scenarios with random motion to demonstrate the localization and tracking capability of the proposed estimation algorithm and ignore optimal robot path planning.\par
			 
			We conduct 50 Monte Carlo simulations and compare the proposed deformable SLAM, rigid SLAM approach and ED node based method. Note that different from the proposed method and rigid SLAM, ED node approach is a visual odometry like approach making it inherently less accurate than the other two methods. Table \ref{Table:monte_carlo_results} shows the comparisons. On the basis of these results we concluded that the proposed deformable formulation outperforms rigid SLAM and ED based approach.\par 
			
			The results are compared by root mean squared errors (RMSE) which quantify the estimation accuracy. Fig. \ref{fig:RMSE_MonteCarlo} is a typical monte carlo simulation showing the RMSE overtime.\par

			\subsection{SLAM in deformable soft-tissues}
			The proposed prior based deformable SLAM is also validated on ex-vivo experiments. In simulation validation step, three different soft-tissue models (heart, liver and lung), which are segmented from a CT scan of a phantom, deform over time. The 3D deforming data are projected into 2D space and we simulate a robot moving inside each soft-tissues. The viewing angle of the robot is $60^{\circ}$. Fig. \ref{fig:trajectory_2models} shows the trajectory of the moving robot as well as the feature positions. The initial state of the feature and robot pose are estimated with traditional visual odometry. Fig. \ref{fig:RMSE_3models} presents the results of the three trajectories in the form of root-mean-square error (RMSE).\par
			
			We also test the dataset on Hamlyn dataset 11 and 12. The camera remains stable (Fig. \ref{fig:staticcamera}) observing two deforming soft-tissues. We track some key points and project them into 2D features to test if the estimated camera pose is stable. Results demonstrates that our algorithm achieves better camera pose (Average error 1.352 mm) than conventional SLAM (Average error 5.473 mm).\par 
			
			These results imply that the proposed priori attributes to ourperforming conventional approaches, which results from the fact that many dataset conforms to the mixture of historical shapes.\par

			\subsection{Observability test}
			\label{secetion:Discussion}
			To gain more insight into observability in the proposed SLAM system, We examine the parameter observability properties by testing Hessian matrix of all the tests. The study of parameter observability to analyze if unique solution of the problem can be found; when parameter observability holds, the Fisher information matrix (FIM) is invertible \cite{bar2004estimation}. We can gain insights about the null space due to the fact that FIM encapsulates all the information available. Section \ref{section:priorbasedSLAM} shows that the null space of the proposed method lies in the deformable parameters $\mathbf{c}$. After testing on all datasets, we find that the Hessian (FIM) has a nullspace of size of $\mathbf{c}$. We also test that Hessian becomes full when $\mathbf{c}$ is fixed. Therefore, even though $\mathbf{c}$ is not fully observable, the robot pose and feature positions are still unique. This test validate our theoretical analysis in Section \ref{observability_test_deformable}.\par 
			
			\begin{table}[!h]
				\caption{Feature estimation accuracies (m) in three models. All the simulation noises (invariances) are set to be 0.1 m.}
				\label{Table:Monte_Carlo_Simulation}
				\begin{center}
					\begin{tabular}{|p{2.0cm}|p{1.4cm}|p{1.4cm}|p{1.4cm}|}
						\hline
						&Heart& Stomach & Lung \\

						\hline
						Estimation error&1.242&2.120 & 3.197\\
						\hline
					\end{tabular}
				\end{center}
				\label{Table:monte_carlo_results}
			\end{table}

			\begin{figure}
				\centering 
				\includegraphics[width=0.5\textwidth]{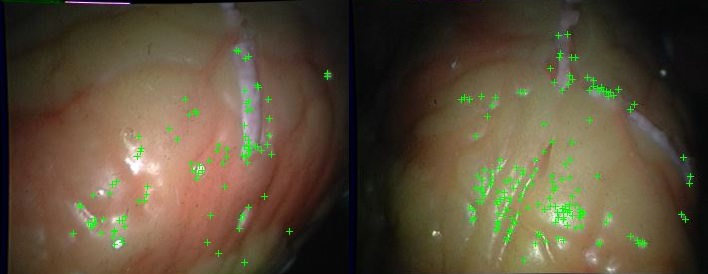}
				\caption{Ground truth dataset from Hamlyn center.} 
				\label{fig:staticcamera}
			\end{figure}

			\section{Conclusion}
			In this paper, our research extends the knowledge of observability analysis into deformable SLAM environment. We perform parameter observability analysis on ED parameterization and prove that in the case of no prior, the global pose is not separable from ED based deformation parameterization. Proofs of the existence multiple solutions are provided for the ED based deformation formulation. The null space in both ED based formulation and rigid SLAM formulation makes the pose estimation not accurate. Based on our discussion, robot pose and deforming environment in SLAM problems are entangled and cannot be estimated without priors. \par 
			To solve this, a new time series priori based algorithm is introduced for localizing robot as well as estimating the deformable environment, when robots operate in dynamic scenario. We prove that the priori is enough to avoid ambiguity of rigid and non-rigid motions of the robot and the environment. The proposed algorithm is validated extensively on Monte Carlo simulations and medical datasets. It significantly outperforms conventional rigid SLAM formulation as well as ED formulation especially in scenario with large and mixture of periodic deformations.

			




			\bibliographystyle{ieeetr}
			\bibliography{reference}   

\begin{thebibliography}{10}

\bibitem{mountney2009dynamic}
P.~Mountney and G.-Z. Yang, ``Dynamic view expansion for minimally invasive
  surgery using simultaneous localization and mapping,'' in {\em 2009 Annual
  International Conference of the IEEE Engineering in Medicine and Biology
  Society}, pp.~1184--1187, IEEE, 2009.

\bibitem{lin2015video}
B.~Lin, Y.~Sun, X.~Qian, {\em et~al.}, ``Video-based 3d reconstruction,
  laparoscope localization and deformation recovery for abdominal minimally
  invasive surgery: a survey,'' {\em The International Journal of Medical
  Robotics and Computer Assisted Surgery}, 2015.

\bibitem{stoyanov2012stereoscopic}
D.~Stoyanov, ``Stereoscopic scene flow for robotic assisted minimally invasive
  surgery,'' in {\em International Conference on Medical Image Computing and
  Computer-Assisted Intervention}, pp.~479--486, Springer, 2012.

\bibitem{haouchine2015monocular}
N.~Haouchine, J.~Dequidt, M.-O. Berger, and S.~Cotin, ``Monocular 3d
  reconstruction and augmentation of elastic surfaces with self-occlusion
  handling,'' {\em IEEE transactions on visualization and computer graphics},
  vol.~21, no.~12, pp.~1363--1376, 2015.

\bibitem{malti2011template}
A.~Malti, A.~Bartoli, and T.~Collins, ``Template-based conformal
  shape-from-motion from registered laparoscopic images.,'' in {\em MIUA},
  vol.~1, p.~6, 2011.

\bibitem{grasa2011ekf}
O.~G. Grasa, J.~Civera, and J.~Montiel, ``Ekf monocular slam with
  relocalization for laparoscopic sequences,'' in {\em Robotics and Automation
  (ICRA), 2011 IEEE International Conference on}, pp.~4816--4821, IEEE, 2011.

\bibitem{lin2013simultaneous}
B.~Lin, A.~Johnson, X.~Qian, J.~Sanchez, and Y.~Sun, ``Simultaneous tracking,
  3d reconstruction and deforming point detection for stereoscope guided
  surgery,'' in {\em Augmented Reality Environments for Medical Imaging and
  Computer-Assisted Interventions}, pp.~35--44, Springer, 2013.

\bibitem{du2015robust}
X.~Du, N.~Clancy, {\em et~al.}, ``Robust surface tracking combining features,
  intensity and illumination compensation,'' {\em International Journal of
  Computer Assisted Radiology and Surgery}, vol.~10, no.~12, pp.~1915--1926,
  2015.

\bibitem{newcombe2011kinectfusion}
R.~A. Newcombe, S.~Izadi, {\em et~al.}, ``Kinectfusion: Real-time dense surface
  mapping and tracking,'' in {\em Mixed and Augmented Reality (ISMAR), 2011
  10th IEEE International Symposium on}, pp.~127--136, IEEE, 2011.

\bibitem{newcombe2015dynamicfusion}
R.~A. Newcombe, D.~Fox, and S.~M. Seitz, ``Dynamicfusion: Reconstruction and
  tracking of non-rigid scenes in real-time,'' in {\em Proceedings of the IEEE
  Conference on Computer Vision and Pattern Recognition}, pp.~343--352, 2015.

\bibitem{innmann2016volumedeform}
M.~Innmann, M.~Zollh{\"o}fer, M.~Nie{\ss}ner, C.~Theobalt, and M.~Stamminger,
  ``Volumedeform: Real-time volumetric non-rigid reconstruction,'' in {\em
  European Conference on Computer Vision}, pp.~362--379, Springer, 2016.

\bibitem{dou2016fusion4d}
M.~Dou, S.~Khamis, {\em et~al.}, ``Fusion4d: real-time performance capture of
  challenging scenes,'' {\em ACM Transactions on Graphics (TOG)}, vol.~35,
  no.~4, p.~114, 2016.

\bibitem{maier2013optical}
L.~Maier-Hein, P.~Mountney, A.~Bartoli, H.~Elhawary, D.~Elson, A.~Groch,
  A.~Kolb, M.~Rodrigues, J.~Sorger, S.~Speidel, {\em et~al.}, ``Optical
  techniques for 3d surface reconstruction in computer-assisted laparoscopic
  surgery,'' {\em Medical image analysis}, vol.~17, no.~8, pp.~974--996, 2013.

\bibitem{sorkine2007rigid}
O.~Sorkine and M.~Alexa, ``As-rigid-as-possible surface modeling,'' in {\em
  Symposium on Geometry Processing}, vol.~4, 2007.

\bibitem{billings2012system}
S.~Billings, N.~Deshmukh, {\em et~al.}, ``System for robot-assisted real-time
  laparoscopic ultrasound elastography,'' in {\em SPIE Medical Imaging},
  pp.~83161--83161, International Society for Optics and Photonics, 2012.

\bibitem{zhang2016autonomous}
L.~Zhang, M.~Ye, {\em et~al.}, ``Autonomous scanning for endomicroscopic
  mosaicing and 3d fusion,'' {\em arXiv preprint arXiv:1604.04137}, 2016.

\bibitem{volumedeformurl}
M.~Innmann, M.~Zollh{\"o}fer, {\em et~al.}, ``Volumedeform: Real-time
  volumetric non-rigid reconstruction - eccv 2016.''
  \url{https://www.youtube.com/watch?v=khthUS7KVY4}, 2016.

\bibitem{uijlings2009real}
J.~R. Uijlings, A.~W. Smeulders, and R.~J. Scha, ``Real-time bag of words,
  approximately,'' in {\em Proceedings of the ACM international Conference on
  Image and Video Retrieval}, p.~6, ACM, 2009.

\bibitem{sumner2007embedded}
R.~W. Sumner, J.~Schmid, and M.~Pauly, ``Embedded deformation for shape
  manipulation,'' {\em ACM Transactions on Graphics (TOG)}, vol.~26, no.~3,
  p.~80, 2007.

\bibitem{giannarou2013probabilistic}
S.~Giannarou, M.~Visentini-Scarzanella, and G.-Z. Yang, ``Probabilistic
  tracking of affine-invariant anisotropic regions,'' {\em IEEE transactions on
  Pattern Analysis and Machine Intelligence}, vol.~35, no.~1, pp.~130--143,
  2013.

\bibitem{kenngott2015openhelp}
H.~Kenngott, J.~W{\"u}nscher, M.~Wagner, {\em et~al.}, ``Openhelp (heidelberg
  laparoscopy phantom): development of an open-source surgical evaluation and
  training tool,'' {\em Surgical Endoscopy}, vol.~29, no.~11, pp.~3338--3347,
  2015.

\bibitem{song20163d}
J.~Song, J.~Wang, L.~Zhao, S.~Huang, and G.~Dissanayake, ``3d shape recovery of
  deformable soft-tissue with computed tomography and depth scan,'' in {\em
  Australasian Conference on Robotics and Automation (ACRA)}, ARAA, 2016.

\end{thebibliography}


\begin{thebibliography}{10}

\bibitem{klein2007parallel}
G.~Klein and D.~Murray, ``Parallel tracking and mapping for small ar
  workspaces,'' in {\em Mixed and Augmented Reality, 2007. ISMAR 2007. 6th IEEE
  and ACM International Symposium on}, pp.~225--234, IEEE, 2007.

\bibitem{newcombe2011kinectfusion}
R.~A. Newcombe, S.~Izadi, {\em et~al.}, ``Kinectfusion: Real-time dense surface
  mapping and tracking,'' in {\em Mixed and Augmented Reality (ISMAR), 2011
  10th IEEE International Symposium on}, pp.~127--136, IEEE, 2011.

\bibitem{engel2014lsd}
J.~Engel, T.~Sch{\"o}ps, and D.~Cremers, ``Lsd-slam: Large-scale direct
  monocular slam,'' in {\em European Conference on Computer Vision},
  pp.~834--849, Springer, 2014.

\bibitem{mur2015orb}
R.~Mur-Artal, J.~M.~M. Montiel, and J.~D. Tardos, ``Orb-slam: a versatile and
  accurate monocular slam system,'' {\em IEEE Transactions on Robotics},
  vol.~31, no.~5, pp.~1147--1163, 2015.

\bibitem{dai2017bundlefusion}
A.~Dai, M.~Nie{\ss}ner, M.~Zollh{\"o}fer, S.~Izadi, and C.~Theobalt,
  ``Bundlefusion: Real-time globally consistent 3d reconstruction using
  on-the-fly surface reintegration,'' {\em ACM Transactions on Graphics (TOG)},
  vol.~36, no.~4, p.~76a, 2017.

\bibitem{grasa2011ekf}
O.~G. Grasa, J.~Civera, and J.~Montiel, ``Ekf monocular slam with
  relocalization for laparoscopic sequences,'' in {\em Robotics and Automation
  (ICRA), 2011 IEEE International Conference on}, pp.~4816--4821, IEEE, 2011.

\bibitem{lin2013simultaneous}
B.~Lin, A.~Johnson, X.~Qian, J.~Sanchez, and Y.~Sun, ``Simultaneous tracking,
  3d reconstruction and deforming point detection for stereoscope guided
  surgery,'' in {\em Augmented Reality Environments for Medical Imaging and
  Computer-Assisted Interventions}, pp.~35--44, Springer, 2013.

\bibitem{mahmoud2016orbslam}
N.~Mahmoud, I.~Cirauqui, A.~Hostettler, C.~Doignon, L.~Soler, J.~Marescaux, and
  J.~Montiel, ``Orbslam-based endoscope tracking and 3d reconstruction,'' in
  {\em International Workshop on Computer-Assisted and Robotic Endoscopy},
  pp.~72--83, Springer, 2016.

\bibitem{mahmoud2017slam}
N.~Mahmoud, A.~Hostettler, T.~Collins, L.~Soler, C.~Doignon, and J.~Montiel,
  ``Slam based quasi dense reconstruction for minimally invasive surgery
  scenes,'' {\em arXiv preprint arXiv:1705.09107}, 2017.

\bibitem{turan2017non}
M.~Turan, Y.~Almalioglu, H.~Araujo, E.~Konukoglu, and M.~Sitti, ``A non-rigid
  map fusion-based direct slam method for endoscopic capsule robots,'' {\em
  International journal of intelligent robotics and applications}, vol.~1,
  no.~4, pp.~399--409, 2017.

\bibitem{chen2018slam}
L.~Chen, W.~Tang, N.~W. John, T.~R. Wan, and J.~J. Zhang, ``Slam-based dense
  surface reconstruction in monocular minimally invasive surgery and its
  application to augmented reality,'' {\em Computer methods and programs in
  biomedicine}, vol.~158, pp.~135--146, 2018.

\bibitem{mahmoud2019live}
N.~Mahmoud, T.~Collins, A.~Hostettler, L.~Soler, C.~Doignon, and J.~M.~M.
  Montiel, ``Live tracking and dense reconstruction for handheld monocular
  endoscopy,'' {\em IEEE transactions on medical imaging}, vol.~38, no.~1,
  pp.~79--89, 2019.

\bibitem{sorkine2007rigid}
O.~Sorkine and M.~Alexa, ``As-rigid-as-possible surface modeling,'' in {\em
  Symposium on Geometry Processing}, vol.~4, pp.~109--116, 2007.

\bibitem{saputra2018visual}
M.~R.~U. Saputra, A.~Markham, and N.~Trigoni, ``Visual slam and structure from
  motion in dynamic environments: A survey,'' {\em ACM Computing Surveys
  (CSUR)}, vol.~51, no.~2, p.~37, 2018.

\bibitem{newcombe2015dynamicfusion}
R.~A. Newcombe, D.~Fox, and S.~M. Seitz, ``Dynamicfusion: Reconstruction and
  tracking of non-rigid scenes in real-time,'' in {\em Proceedings of the IEEE
  Conference on Computer Vision and Pattern Recognition}, pp.~343--352, 2015.

\bibitem{innmann2016volumedeform}
M.~Innmann, M.~Zollh{\"o}fer, M.~Nie{\ss}ner, C.~Theobalt, and M.~Stamminger,
  ``Volumedeform: Real-time volumetric non-rigid reconstruction,'' in {\em
  European Conference on Computer Vision}, pp.~362--379, Springer, 2016.

\bibitem{guo2017real}
K.~Guo, F.~Xu, T.~Yu, X.~Liu, Q.~Dai, and Y.~Liu, ``Real-time geometry, albedo,
  and motion reconstruction using a single rgb-d camera,'' {\em ACM
  Transactions on Graphics (TOG)}, vol.~36, no.~3, p.~32, 2017.

\bibitem{dou2016fusion4d}
M.~Dou, S.~Khamis, {\em et~al.}, ``Fusion4d: real-time performance capture of
  challenging scenes,'' {\em ACM Transactions on Graphics (TOG)}, vol.~35,
  no.~4, p.~114, 2016.

\bibitem{song2018dynamic}
J.~Song, J.~Wang, L.~Zhao, S.~Huang, and G.~Dissanayake, ``Dynamic
  reconstruction of deformable soft-tissue with stereo scope in minimal
  invasive surgery,'' {\em IEEE Robotics and Automation Letters}, vol.~3,
  no.~1, pp.~155--162, 2018.

\bibitem{song2018mis}
J.~Song, J.~Wang, L.~Zhao, S.~Huang, and G.~Dissanayake, ``Mis-slam: Real-time
  large scale dense deformable slam system in minimal invasive surgery based on
  heterogeneous computing,'' {\em arXiv preprint arXiv:1803.02009}, 2018.

\bibitem{agudo2011fem}
A.~Agudo, B.~Calvo, and J.~Montiel, ``Fem models to code non-rigid ekf
  monocular slam,'' in {\em Computer Vision Workshops (ICCV Workshops), 2011
  IEEE International Conference on}, pp.~1586--1593, IEEE, 2011.

\bibitem{agudo20123d}
A.~Agudo, B.~Calvo, and J.~Montiel, ``3d reconstruction of non-rigid surfaces
  in real-time using wedge elements,'' in {\em European Conference on Computer
  Vision}, pp.~113--122, Springer, 2012.

\bibitem{bar2004estimation}
Y.~Bar-Shalom, X.~R. Li, and T.~Kirubarajan, {\em Estimation with applications
  to tracking and navigation: theory algorithms and software}.
\newblock John Wiley \& Sons, 2004.

\bibitem{yip1981solvability}
E.~Yip and R.~Sincovec, ``Solvability, controllability, and observability of
  continuous descriptor systems,'' {\em IEEE Transactions on Automatic
  Control}, vol.~26, no.~3, pp.~702--707, 1981.

\bibitem{thrun2005probabilistic}
S.~Thrun, W.~Burgard, and D.~Fox, {\em Probabilistic robotics}.
\newblock MIT press, 2005.

\bibitem{sumner2007embedded}
R.~W. Sumner, J.~Schmid, and M.~Pauly, ``Embedded deformation for shape
  manipulation,'' {\em ACM Transactions on Graphics (TOG)}, vol.~26, no.~3,
  p.~80, 2007.

\bibitem{agudo2017force}
A.~Agudo and F.~Moreno-Noguer, ``Force-based representation for non-rigid shape
  and elastic model estimation,'' {\em IEEE transactions on pattern analysis
  and machine intelligence}, vol.~40, no.~9, pp.~2137--2150, 2017.

\bibitem{agudo2018robust}
A.~Agudo and F.~Moreno-Noguer, ``Robust spatio-temporal clustering and
  reconstruction of multiple deformable bodies,'' {\em IEEE transactions on
  pattern analysis and machine intelligence}, vol.~41, no.~4, pp.~971--984,
  2018.

\bibitem{garg2013dense}
R.~Garg, A.~Roussos, and L.~Agapito, ``Dense variational reconstruction of
  non-rigid surfaces from monocular video,'' in {\em Proceedings of the IEEE
  Conference on Computer Vision and Pattern Recognition}, pp.~1272--1279, 2013.

\bibitem{valmadre2012general}
J.~Valmadre and S.~Lucey, ``General trajectory prior for non-rigid
  reconstruction,'' in {\em 2012 IEEE Conference on Computer Vision and Pattern
  Recognition}, pp.~1394--1401, IEEE, 2012.

\bibitem{simon2014separable}
T.~Simon, J.~Valmadre, I.~Matthews, and Y.~Sheikh, ``Separable spatiotemporal
  priors for convex reconstruction of time-varying 3d point clouds,'' in {\em
  European Conference on Computer Vision}, pp.~204--219, Springer, 2014.

\bibitem{absil2007trust}
P.-A. Absil, C.~G. Baker, and K.~A. Gallivan, ``Trust-region methods on
  riemannian manifolds,'' {\em Foundations of Computational Mathematics},
  vol.~7, no.~3, pp.~303--330, 2007.

\bibitem{zhang2017convergence}
T.~Zhang, K.~Wu, J.~Song, S.~Huang, and G.~Dissanayake, ``Convergence and
  consistency analysis for a 3-d invariant-ekf slam,'' {\em IEEE Robotics and
  Automation Letters}, vol.~2, no.~2, pp.~733--740, 2017.

\end{thebibliography}

		\end{document}